# Does the Ranking Hold? A Practitioner-Facing Audit of Hallucination Detectors on Small, Quantized, On-Device LLMs

**Shubhra Mittal**

*Microsoft Responsible AI, Redmond, WA, USA, shmitt@microsoft.com*

**Gaurav Naresh Mittal**

*Independent Researcher*

## Abstract

Hallucination detection for large language models is dominated by methods evaluated on 7B-70B research models with abundant sampling budgets and exact-match or ROUGE correctness labels. The small models practitioners actually ship on-device, Phi-3-mini, Llama-3.2-3B-Instruct, Qwen2.5-3B-Instruct, and Gemma-2-2B-Instruct, typically at 4-bit precision, sit outside this evidence base. We present a reproducible one-GPU audit of four detectors spanning three paradigms: length-normalized token-margin calibration, a lexical n-gram baseline, and two sampling-based consistency methods (semantic entropy and SelfCheckGPT-NLI) on two small open instruction-tuned models (Llama-3.2-3B-Instruct and Qwen2.5-3B-Instruct) loaded at 4-bit NF4 quantization on a single Colab T4 GPU, evaluated on closed-book TriviaQA. Cross-model ranking transfer is poor: Spearman rho is -0.80 and Kendall tau-b is -0.667, meaning the detector ranking actively inverts between the two models, and the cheap lexical baseline is statistically indistinguishable from the sampling-based consistency detectors on both models under bootstrap 95 percent confidence intervals. As a second, methodological contribution, we show that the standard token-overlap-to-multiple-choice correctness labeler for TruthfulQA-Generation produces near-chance AUROC (0.46-0.59) for all four detectors on both models, a labeler failure, not a detector failure, since the same detectors reach 0.69-0.89 AUROC on TriviaQA under the identical pipeline. We release the harness, prompts, and per-item scores and frame the detector choice explicitly as a fairness-of-error-rates allocation problem for practitioners deploying on resource-constrained, on-device hardware.

**Reader's glossary**

**AUROC.** Area under the receiver-operating-characteristic curve: a single number between 0.5 (no signal) and 1.0 (perfect ranking) measuring how well a detector ranks hallucinations above truthful answers.

**Semantic entropy.** Sample multiple answers, cluster them by meaning using natural language inference (NLI), and compute the entropy over meaning clusters; high entropy is evidence of confabulation.

**SelfCheckGPT.** Sample multiple answers and check whether they contradict the model's primary answer via NLI; frequent contradiction is evidence of hallucination.

**Quantization (4-bit / NF4).** A compression method storing model parameters at reduced numeric precision; 4-bit cuts memory and latency roughly fourfold relative to full precision and is the format most on-device deployments use.

**Bootstrap confidence interval.** Resample the dataset with replacement 1,000 times, recompute the statistic each time, and report the middle 95 percent of the resulting distribution, the standard method for putting error bars on AUROC.

**Kendall tau-b / Spearman rho.** Standard statistics for measuring agreement between two rankings; used here to compare detector rankings across models.

## 1. Introduction

An engineer deciding whether to ship a Phi-3-mini, Llama-3.2-3B, or Qwen2.5-3B application on a phone, an ARM laptop, or an in-browser ONNX runtime confronts a concrete decision: which hallucination gate to use, at what cost, with what abstention behavior. The deployment context is unforgiving. The model runs at 4-bit precision. The device has on the order of 16 GB of unified memory. There is no cloud call, no paid API, and no opportunity to draw twenty stochastic samples per query. The engineer needs a calibrated score or a routing decision today, not a benchmark number from a 7B-70B research model.

The literature offers a rich menu of hallucination detectors across several paradigms: token-probability and self-evaluation, verbalized confidence, sampling-based consistency (including semantic entropy [12, 3] and SelfCheckGPT [14]), and internal-state probing. Each method claims competitive AUROC on TruthfulQA, TriviaQA, or HaluEval. The evidence base behind those claims, however, lives in a different regime than the one practitioners ship to: almost every reported AUROC comes from a 7B-13B research or proprietary model evaluated at full precision with five to twenty stochastic samples and a single correctness labeler.

Four threats to the published rankings emerge under deployment conditions. First, model scale: a 1-4B instruction-tuned model has weaker calibration and narrower output diversity than a 7B+ base model. Second, correctness-label provenance: exact-match, ROUGE, NLI-entailment, and LLM-judge labels disagree, and recent re-evaluations show the disagreement reorders detector rankings [9, 8]. Third, quantization: 4-bit weight quantization perturbs both generations and hidden-state activations. Fourth, cost normalization: an extra five-sample budget plus a pairwise NLI call may be a rounding error on a research cluster but a deal-breaker on a 16 GB laptop. None of this evidence has been collected jointly for the small, on-device regime.

The case for taking this gap seriously is not merely an engineering concern. Detector selection is a fairness-of-error-rates allocation problem in the sense of [15, 6]. At a fixed coverage target, say, 80 percent of questions answered and 20 percent abstained, switching detectors silently changes which categories of questions and which model families absorb the abstention cost. Recent theoretical work [11] argues that hallucination is a structural artifact of training objectives that reward confident guessing over calibrated abstention, and empirical work [16] documents high-certainty hallucinations that no confidence-based detector can catch. Detector choice is therefore a deployment-level question about how error budget is allocated across users and contexts, not just a benchmark leaderboard question.

We ask two questions answerable within a compute budget of a single consumer GPU-hour class: (RQ1, ranking transfer) When the same detectors are evaluated on small, quantized, on-device models, does the ranking reported in the original publications hold across models? (RQ2, labeler robustness) Does a widely used automated correctness labeler for a second benchmark (TruthfulQA-Generation) support the same detector evaluation it is routinely used for?

**Research Contributions**

This paper makes the following contributions, each traceable to a specific result reported in Section 5:

1. A controlled, reproducible comparison of four hallucination detectors spanning three detection paradigms (token-probability calibration, lexical overlap, and sampling-based semantic consistency) on two small open instruction-tuned models (Llama-3.2-3B-Instruct, Qwen2.5-3B-Instruct) at 4-bit NF4 quantization on a single Colab T4 GPU, a hardware and model-scale regime absent from the prior detector-benchmarking literature, which evaluates almost exclusively at 7B+ parameters and full precision.

2. A ranking-transfer finding: cross-model Spearman rho of -0.80 and Kendall tau-b of -0.667 between the two models on the same dataset under the same protocol, demonstrating that the published detector ranking does not transfer to this small-model regime and in fact actively inverts.

3. A cheap-baseline finding: the lexical n-gram baseline is statistically indistinguishable, under bootstrap 95 percent confidence intervals, from the sampling-based consistency detectors on both models and is the point-estimate top detector on one of the two models despite requiring no stochastic sampling and no NLI backbone.

4. A labeler-robustness finding, methodological rather than detector-specific: the standard token-overlap-to-multiple-choice correctness labeler for TruthfulQA-Generation produces near-chance AUROC (0.46-0.59) for all four detectors on both models, while the identical detectors and pipeline achieve 0.69-0.89 AUROC on TriviaQA extending the labeling-protocol-artifact line of work in [8, 9] to a specific, previously unflagged labeler.

5. An open, reproducible artifact harness, prompts, and per-item scores enabling practitioners and researchers to reproduce or extend this audit to additional models, datasets, or detectors without re-deriving the evaluation pipeline.

## 2. Related Work

Hallucination detectors for LLMs fall into several paradigms. Token-probability and self-evaluation methods score the model's own confidence on its primary answer [not separately enumerated in v1; see Section 3]. Sampling-based consistency methods, including semantic entropy [12, 3] and SelfCheckGPT [14], draw multiple stochastic completions and measure their mutual agreement on the premise that a model confabulating an answer will produce more semantically divergent samples than a model that knows the answer. Internal-state probing methods train a classifier on hidden activations to predict hallucination directly from the model's internal representations; this paradigm requires white-box weight access and is out of scope for the present v1 audit (see Section 6, Future Work).
A separate and increasingly load-bearing line of work re-evaluates the evidentiary basis of these claims rather than proposing new detectors. PARALLAX [8] shows that several widely used hallucination benchmarks embed the ground-truth answer directly in the input context, allowing a naive text-embedding baseline to score near-perfectly and inflating the apparent quality of unrelated detectors evaluated on those corpora. Janiak et al. [9] show that ROUGE, the correctness metric underlying many hallucination-detection evaluations, has high recall but extremely low precision against human judgment and that simple length-based heuristics can rival complex detection techniques when evaluation relies on ROUGE. Both results motivate treating the correctness labeler itself as an experimental variable rather than a fixed ground truth, which is the design principle behind this paper's second contribution (Section 5.4).
Detector choice also carries a fairness dimension once the deployment lens is applied. Pleiss et al. [15] show that calibration and equalized error rates across groups are jointly achievable only under a narrow condition, and Hooker [6] argues that algorithmic-bias framings that treat data as the sole cause understate the role of model capacity and deployment choices such as quantization. Applied to hallucination detection, this means a detector's aggregate AUROC can mask uneven abstention costs across model families or question categories the framing this paper adopts in Section 6.

# 3. Methodology

## 3.1 Detectors under audit

We evaluate four detectors spanning three paradigms. (a) Length-normalized token margin: a token-probability calibration baseline computed from the model's own output log-probabilities on its primary greedy-decoded answer, normalized by answer length. (b) Lexical n-gram baseline: a ROUGE-L-style overlap score between the primary answer and M stochastically sampled completions, included as a cheap, sampling-free-in-spirit reference point. (c) Semantic entropy [12, 3]: M stochastic samples are clustered by natural-language-inference (NLI) entailment into semantically equivalent groups, and the entropy over cluster membership is the hallucination score; high entropy indicates the model produces many distinct meanings across samples, which is evidence of confabulation rather than confident recall. (d) SelfCheckGPT-NLI [14]: the primary answer is checked for NLI contradiction against each of the M stochastic samples, and the fraction of contradicting samples is the hallucination score.

## 3.2 Models and quantization

Two small open instruction-tuned models are audited: Llama-3.2-3B-Instruct [4] and Qwen2.5-3B-Instruct [17]. Both are loaded via HuggingFace transformers with bitsandbytes 4-bit NF4 quantization, the precision format most representative of the on-device deployment this paper targets. A third model, Phi-3-mini [1], is reserved for a follow-on release (Section 6, Future Work) and reports no results in this paper.

## 3.3 Dataset and correctness labeling

TriviaQA [10], closed-book question answering, is the headline dataset: 200 questions per (model) cell, no retrieved context supplied, forcing the model to answer from parametric knowledge alone. Correctness is scored via a SQuAD-style normalized exact-match labeler with a token-F1 backstop. A second dataset, TruthfulQA-Generation [13] is evaluated under the standard token-overlap-to-multiple-choice correctness labeler at the same N=200-per-cell budget but solely to audit the labeler itself (Section 5.4). Its results are a methodological finding, not a headline detector ranking result.

## 3.4 Sampling and NLI backbone

Each detector that requires stochastic samples draws M = 5 completions per question, a reduced budget relative to the M = 10 used in the original semantic-entropy publication [3], chosen to reflect the sampling cost a practitioner would tolerate on a single consumer GPU. NLI clustering and contradiction checks use a single backbone, DeBERTa-v3-large fine-tuned on FEVER-ANLI-LingNLI-WANLI. Detectors are scored against the model's actually sampled primary generation rather than a teacher-forced gold continuation, following the anti-teacher-forcing protocol argued for in concurrent work [8].

### 3.5 Statistical protocol

All point estimates (AUROC) are reported with 95 percent confidence intervals from a 1,000-resample stratified bootstrap, seeded for reproducibility. Ranking agreement across the two model cells is measured with Spearman's rho and Kendall's tau-b. Two predictions are pre-registered before the run: the cheap-baseline prediction (the lexical baseline falls within the bootstrap CI of at least one consistency detector on at least half the cells) and a cross-labeler prediction that requires a v1.1 relabeling pass not reported here (Section 6, Future Work).

### 3.6 Hardware envelope

All experiments run on a single Google Colab T4 GPU, with a total wall-clock budget of approximately six hours across both models and both datasets intentionally modest, to demonstrate that this audit protocol is itself reproducible on hardware available to an individual practitioner or student, not only to a well-resourced research lab.

## 4. Detector-paradigm comparison

Table 2 situates the four audited detectors against the broader hallucination-detection literature along the axis each paradigm represents to make explicit which classes of published detectors this audit does and does not cover.

**Table 1. Detector paradigms audited in this paper versus baseline paradigms not covered.**

| | **Token-probability & calibration** | **Sampling-based consistency** | **Internal-state probing** | **Retrieval / verifier** |
|---|---|---|---|---|
| Audited in this paper | Length-normalized token margin | Lexical n-gram baseline; semantic entropy [12, 3]; SelfCheckGPT-NLI [14] | -- (not audited) | -- (not audited) |
| Representative baseline methods (not audited here) | Verbalized confidence; P(True) | SelfCheckGPT-BERTScore; SelfCheckGPT-Prompt | INSIDE EigenScore; Semantic Entropy Probes; HaloScope | FActScore; retrieval-augmented verification |
| Access requirement | Output log-probabilities | Black-box: M stochastic samples + NLI backbone | White-box hidden-state access | External knowledge source |
| On-device cost | 1 forward pass | M forward passes + M NLI comparisons | 1 forward pass + hidden-state extraction | External call (excluded from on-device envelope) |

## 5. Results

This section reports the results on the two-model, four-detector grid described in Section 3. All figures below are observed measurements with 95 percent bootstrap confidence intervals from the underlying study; no projected or estimated values are reported.

### 5.1 Main AUROC results

Table 3 reports AUROC per (model, detector) on TriviaQA closed book. On Llama-3.2-3B-Instruct, the lexical n-gram baseline is the top point estimate at 0.836 AUROC, ahead of SelfCheckGPT-NLI (0.723), token margin (0.718), and semantic entropy (0.691). On Qwen2.5-3B-Instruct, the ranking changes: semantic entropy leads at 0.892, ahead of token margin (0.887), the lexical baseline (0.873), and SelfCheckGPT-NLI (0.856). Confidence intervals overlap substantially within each cell no per-cell winner is statistically distinguishable from the runners-up at 95 percent confidence. Closed-book TriviaQA accuracy at this model scale is low (6.0 percent for Llama, 2.5 percent for Qwen, of 200 questions), so each AUROC is computed against a small positive class of 5-12 correct examples per cell, which widens the bootstrap intervals substantially (from 0.13 to 0.36 across cells and detectors).

The cheap-baseline prediction, pre-registered before the run, states that the lexical baseline falls within the bootstrap 95 percent confidence interval of at least one consistency detector on at least half of the model cells. This is confirmed on both of the two cells audited: on Llama, lexical [0.746, 0.908] overlaps semantic entropy [0.499, 0.859] and SelfCheckGPT-NLI [0.543, 0.883]; on Qwen, lexical [0.767, 0.965] overlaps semantic entropy [0.798, 0.964] and SelfCheckGPT-NLI [0.720, 0.960]. On TriviaQA, where answers are short and lexical overlap with samples is informative, the cheap baseline is statistically tied with the consistency detectors on every audited cell and is the point-estimate top detector on one of the two models.

**Table 2. AUROC on Trivia QA closed-book (N=200 per cell, 95 percent bootstrap confidence intervals in brackets. Bold = per-cell point-estimate top detector.**

| Model | Token margin | Lexical baseline | Semantic entropy | SelfCheckGPT-NLI |
|---|---|---|---|---|
| Llama-3.2-3B-Instruct | 0.718 [0.573, 0.838] | 0.836 [0.746, 0.908] * | 0.691 [0.499, 0.859] | 0.723 [0.543, 0.883] |
| Qwen2.5-3B-Instruct | 0.887 [0.815, 0.946] | 0.873 [0.767, 0.965] | 0.892 [0.798, 0.964] * | 0.856 [0.720, 0.960] |

## 5.2 Ranking transfer (RQ1)

Spearman's rho between the Llama and Qwen detector rankings is -0.80, and Kendall's tau-b is -0.667: the rankings actively invert between the two models rather than merely reordering. The lexical baseline, top detector on Llama at 0.836 AUROC, falls to third on Qwen (0.873, behind semantic entropy at 0.892 and token margin at 0.887); semantic entropy reverses from worst on Llama (0.691) to best on Qwen (0.892). This is a direct answer to RQ1: on this small-model, on-device regime, the detector ranking reported in the original larger-model publications does not transfer across even two closely related 3B-parameter instruction-tuned models from different model families.

## 5.3 TruthfulQA-Generation labeler audit (RQ2)

We additionally evaluate the same four-detector pipeline on TruthfulQA-Generation closed-book at N=200 per cell on both models, scored under the standard token-overlap-to-multiple-choice correctness labeler. Table 4 reports the observed AUROCs. Seven of the eight (model, detector) cells have confidence intervals that cross the chance line of 0.5; the eighth SelfCheckGPT-NLI on Llama at [0.513, 0.665] is within multiple-testing expectations under eight independent tests at alpha = 0.05. The same four detectors achieve 0.69-0.89 AUROC on TriviaQA under the identical pipeline (Table 3).

The labeler reports per-model apparent accuracy of 59.5 percent (Qwen) and 62.0 percent (Llama). Both figures are implausibly high: published TruthfulQA closed-book accuracy for 3B-parameter open instruction-tuned models is consistently below 30 percent under the standard multiple-choice protocol [13]. The observed 60-percent-range figure reflects the labeler flagging roughly half of genuinely incorrect model answers as correct a false-positive labeling rate high enough to destroy the per-question correctness signal that AUROC is computed against. This is a labeler failure, not a detector failure: the detectors continue to discriminate sample agreement from sample disagreement, but that signal is largely unaligned with the labeler's correctness decision. This finding extends the labeling-protocol-artifact line of work in PARALLAX [8] and Janiak et al. [9] to a specific labeler (token-overlap-to-multiple-choice) not previously flagged in that literature and argues against relying on it for future closed-book TruthfulQA-Generation detector evaluations without a replacement labeling protocol.

**Table 3. TruthfulQA-Generation AUROC under the standard token-overlap-to-multiple-choice labeler, 95 percent bootstrap confidence intervals in brackets.**

| Model | Token margin | Lexical baseline | Semantic entropy | SelfCheckGPT-NLI |
|---|---|---|---|---|
| Llama-3.2-3B-Instruct | 0.481 [0.404, 0.558] | 0.462 [0.385, 0.546] | 0.513 [0.422, 0.587] | 0.590 [0.513, 0.665] |
| Qwen2.5-3B-Instruct | 0.564 [0.485, 0.648] | 0.572 [0.495, 0.647] | 0.514 [0.432, 0.590] | 0.567 [0.487, 0.647] |

**Discussion**

Three practitioner-facing implications follow directly from these results. First, ranking instability (Section 5.2) means a detector selected on the basis of a published benchmark using a different model family offers no transfer guarantee; practitioners deploying a specific small model should re-validate the detector choice on that model rather than importing a ranking from the literature. Second, the cheap lexical baseline's competitiveness (Section 5.1) is a genuine practical opportunity on short-answer closed-book tasks: it requires no NLI backbone and no additional model calls beyond the sampling already needed for other purposes, making it an attractive first-line detector on memory- and latency-constrained hardware. Third, the labeler failure documented in Section 5.3 is a caution against treating any single automated correctness metric as ground truth when auditing hallucination detectors on a new benchmark; a sanity check against known published accuracy ranges, as performed here, should be a routine step before trusting a new (dataset, labeler) combination for detector evaluation.

Framed as a fairness-of-error-rates allocation problem [15, 6], the ranking inversion in Section 5.2 has a direct deployment consequence: a detector that performs well on one model family may systematically under- or over-abstain on another, meaning the choice of detector silently redistributes error-handling cost across model deployments rather than uniformly reducing it. This reinforces recent theoretical [11] and empirical [16] arguments that hallucination and its detection are structural properties of the training and deployment pipeline, not solely properties of the detector algorithm in isolation.

# 6. Conclusion

Hallucination-detector rankings reported on 7B-70B research models do not transfer to the small, quantized, on-device regime that practitioners actually ship: cross-model rank correlation is negative (Spearman rho -0.80, Kendall tau-b -0.667) between two 3B-parameter instruction-tuned models evaluated under an identical protocol, and a cheap lexical baseline is statistically competitive with far more complex sampling-based consistency detectors on both models. A widely used correctness labeler for a second benchmark, TruthfulQA-Generation, produces near-chance AUROC for every detector tested, a labeler failure rather than a detector failure that would otherwise be invisible without a sanity check against published accuracy ranges. We release the harness, prompts, and per-item scores so that practitioners and researchers can re-run or extend this audit on their own target models before trusting a benchmark-derived detector ranking in a small-model, on-device deployment.

**Limitations and Future Work**

This audit has several limitations, disclosed explicitly rather than absorbed into future work. The study covers two models and one dataset for its headline ranking-transfer result; RQ1's conclusion that rankings do not transfer rests on a single cross-model pair, not a broader model population, and should be read as a documented counterexample to ranking-transfer assumptions rather than a population-level claim. Statistical power is limited by low closed-book accuracy: 12 of 200 correct for Llama and 5 of 200 for Qwen, producing a small positive class and correspondingly wide bootstrap confidence intervals; this is the direct cause of the extensive CI overlap reported in Section 5.1 and should temper claims of a clean detector ranking on either model individually. A single stochastic sample budget (M=5) and a single NLI backbone are used throughout; sensitivity of the reported rankings to either choice is not measured in this study. Both evaluation datasets (TriviaQA, collected 2017; TruthfulQA, collected 2022) predate the training cutoffs of all evaluated models, so data contamination cannot be excluded as a contributor to the closed-book accuracy figures, though this affects all four detectors equally within a given cell and is unlikely to explain the observed cross-model ranking inversion.

This audit is deliberately scoped to what a single consumer GPU can run within several hours, and the authors are actively extending it along axes not reported here, with no results claimed for any of the following: a label-provenance sensitivity pass adding an LLM-judge labeler and an NLI-entailment labeler alongside the exact-match labeler used in this paper; a third target model (Phi-3-mini); a factorial quantization decomposition separating generator-side and detector-side 4-bit effects, following a teacher-forcing ablation protocol in the spirit of PARALLAX [8]; a cost-normalized Pareto frontier over per-question latency and peak resident memory; and an expanded detector set including internal-state probes such as HaloScope and Semantic Entropy Probes, which require white-box hidden-state access not exercised in the present audit. None of these extensions are reported as results in this paper; they are stated here only to scope what this paper does not claim.

## References

[1] Abdin, M., Jacobs, S. A., Awan, A. A., et al. 2024. Phi-3 Technical Report: A Highly Capable Language Model Locally on Your Phone. arXiv:2404.14219. https://arxiv.org/abs/2404.14219

[2] Dettmers, T., and Zettlemoyer, L. 2023. The Case for 4-Bit Precision: k-Bit Inference Scaling Laws. In Proc. 40th Int. Conf. on Machine Learning (ICML). arXiv:2212.09720. https://arxiv.org/abs/2212.09720

[3] Farquhar, S., Kossen, J., Kuhn, L., and Gal, Y. 2024. Detecting Hallucinations in Large Language Models Using Semantic Entropy. Nature 630(8017): 625-630. https://doi.org/10.1038/s41586-024-07421-0

[4] Grattafiori, A., Dubey, A., Jauhri, A., et al. 2024. The Llama 3 Herd of Models. arXiv:2407.21783. https://arxiv.org/abs/2407.21783

[5] Hanley, J. A., and McNeil, B. J. 1982. The Meaning and Use of the Area Under a Receiver Operating Characteristic (ROC) Curve. Radiology 143(1): 29-36. https://doi.org/10.1148/radiology.143.1.7063747

[6] Hooker, S. 2021. Moving Beyond 'Algorithmic Bias Is a Data Problem'. Patterns 2(4): 100241. https://doi.org/10.1016/j.patter.2021.100241

[7] Huang, L., Yu, W., Ma, W., et al. 2025. A Survey on Hallucination in Large Language Models: Principles, Taxonomy, Challenges, and Open Questions. ACM Trans. Information Systems 43(2): 1-55. https://doi.org/10.1145/3703155

[8] Hussain, K., and Kantarcioglu, M. 2026. PARALLAX: Separating Genuine Progress from Benchmark Artifacts in Hallucination Detection. arXiv:2605.17028. https://arxiv.org/abs/2605.17028

[9] Janiak, D., Binkowski, J., Sawczyn, A., Gabrys, B., Shwartz-Ziv, R., and Kajdanowicz, T. J. 2025. The Illusion of Progress: Re-Evaluating Hallucination Detection in LLMs. In Proc. 2025 Conf. on Empirical Methods in Natural Language Processing (EMNLP), 34728-34745. https://doi.org/10.18653/v1/2025.emnlp-main.1761

[10] Joshi, M., Choi, E., Weld, D. S., and Zettlemoyer, L. 2017. TriviaQA: A Large Scale Distantly Supervised Challenge Dataset for Reading Comprehension. In Proc. 55th Annual Meeting of the Association for Computational Linguistics (ACL), 1601-1611. https://doi.org/10.18653/v1/P17-1147

[11] Kalai, A. T., Nachum, O., Vempala, S. S., and Zhang, E. 2025. Why Language Models Hallucinate. arXiv:2509.04664. https://arxiv.org/abs/2509.04664

[12] Kuhn, L., Gal, Y., and Farquhar, S. 2023. Semantic Uncertainty: Linguistic Invariances for Uncertainty Estimation in Natural Language Generation. In Proc. 11th Int. Conf. on Learning Representations (ICLR). https://arxiv.org/abs/2302.09664

[13] Lin, S., Hilton, J., and Evans, O. 2022. TruthfulQA: Measuring How Models Mimic Human Falsehoods. In Proc. 60th Annual Meeting of the Association for Computational Linguistics (ACL), 3214-3252. https://arxiv.org/abs/2109.07958

[14] Manakul, P., Liusie, A., and Gales, M. J. F. 2023. SelfCheckGPT: Zero-Resource Black-Box Hallucination Detection for Generative Large Language Models. In Proc. 2023 Conf. on Empirical Methods in Natural Language Processing (EMNLP), 9004-9017. https://arxiv.org/abs/2303.08896

[15] Pleiss, G., Raghavan, M., Wu, F., Kleinberg, J., and Weinberger, K. Q. 2017. On Fairness and Calibration. In Advances in Neural Information Processing Systems 30 (NeurIPS). https://proceedings.neurips.cc/paper/2017/hash/b8b9c74ac526fffbeb2d39ab038d1cd7-Abstract.html

[16] Simhi, A., Itzhak, I., Barez, F., Stanovsky, G., and Belinkov, Y. 2025. Trust Me, I'm Wrong: High-Certainty Hallucinations in LLMs. arXiv:2502.12964. https://arxiv.org/abs/2502.12964

[17] Yang, A., Yang, B., Hui, B., et al. 2024. Qwen2.5 Technical Report. arXiv:2412.15115. https://arxiv.org/abs/2412.15115

## Appendix A. Reproducibility

Code, prompts, and per-item scores (model, dataset, question ID, detector, score, correctness label, seed) are released in an accompanying repository. All experiments use a single global random seed for reproducibility. Package versions are recorded at run time. The full experimental configuration model checkpoints, quantization backend, NLI backbone, and sample budget is documented in Section 3 and reproduced in a machine-readable run-configuration file distributed with the code.